%% file: main.tex
\documentclass[9pt,journal,compsoc]{IEEEtran}

\pdfoutput=1

\ifCLASSOPTIONcompsoc
  \usepackage[nocompress]{cite}
\else
  \usepackage{cite}
\fi

\ifCLASSINFOpdf
   \usepackage[pdftex]{graphicx}
\else
   \usepackage[dvips]{graphicx}
\fi

\usepackage{amsmath,amssymb,amsfonts}
\usepackage{algorithmic}
\usepackage{bm}
\usepackage{multirow}
\usepackage{stfloats}
\usepackage{textcomp}
\usepackage{xcolor}
\usepackage[nocompress]{cite}
\usepackage[top=0.60in,bottom=0.60in,left=0.52in,right=0.52in]{geometry}
\usepackage{fancyhdr}
\usepackage[normalem]{ulem}
\usepackage[hyphens]{url}
\usepackage{hyperref}
\usepackage{color}
\usepackage{soul}
\usepackage{diagbox}
\usepackage{enumitem}
\usepackage{relsize}
\usepackage[square, numbers]{natbib}

\usepackage[labelfont={scriptsize},textfont={scriptsize}]{caption}
\usepackage{multirow}
\usepackage{rotating}
\usepackage{array}
\usepackage{pifont}
\usepackage{adjustbox}

\iftrue
\usepackage{titlesec}
\titlespacing\section{0pt}{3pt plus 1pt minus 1pt}{0pt plus 1pt minus 1pt}
\titlespacing\subsection{0pt}{3pt plus 1pt minus 1pt}{0pt plus 1pt minus 1pt}
\titlespacing\subsubsection{0pt}{3pt plus 1pt minus 1pt}{2pt plus 1pt minus 1pt}
\fi

\begin{document}

\title{Architectural Implications of GNN Aggregation\\ Programming Abstractions}

\author{Yingjie~Qi,
        Jianlei~Yang,
        Ao~Zhou,
        Tong~Qiao,
        and~Chunming~Hu
\IEEEcompsocitemizethanks{
\IEEEcompsocthanksitem This work is supported by National Natural Science Foundation of China (Grant No. 61602022).
\IEEEcompsocthanksitem Corresponding authors are Jianlei Yang, Email: \url{jianlei@buaa.edu.cn}.
\IEEEcompsocthanksitem Y. Qi, J. Yang, A. Zhou, T. Qiao and C. Hu are with School of Computer Science and Engineering, Beihang University, Beijing, China.
}
}

\IEEEtitleabstractindextext{%
\begin{abstract}
Graph neural networks (GNNs) have gained significant popularity due to the powerful capability to extract useful representations from graph data.
As the need for efficient GNN computation intensifies, a variety of programming abstractions designed for optimizing GNN Aggregation have emerged to facilitate acceleration.
However, there is no comprehensive evaluation and analysis upon existing abstractions, thus no clear consensus on which approach is better.
In this letter, we classify existing programming abstractions for GNN Aggregation by the dimension of data organization and propagation method.
By constructing these abstractions on a state-of-the-art GNN library, we perform a thorough and detailed characterization study to compare their performance and efficiency, and provide several insights on future GNN acceleration based on our analysis.
\end{abstract}

\begin{IEEEkeywords}
Graph Neural Networks (GNNs), Characterization, Programming Abstractions, Execution Patterns.
\end{IEEEkeywords}}

\maketitle

\IEEEdisplaynontitleabstractindextext

\IEEEpeerreviewmaketitle

\input{main_text/1-Introduction}
\input{main_text/2-Preliminaries.tex}

\input{main_text/3-Methodology.tex}
\input{main_text/4-Result-Analysis.tex}
\input{main_text/5-Discussion.tex}

\input{main_text/6-Conclusion.tex}

\ifCLASSOPTIONcaptionsoff
  \newpage
\fi

{
\scriptsize
\bibliographystyle{IEEEtran}
\bibliography{./ref/ref}
}

\end{document}

%% file: main_text/1-Introduction.tex
\IEEEraisesectionheading{\section{Introduction}\label{sec:introduction}}

\IEEEPARstart{I}{n} recent years, Graph Neural Networks (GNNs) have flourished in a variety of machine learning scenarios, due to their capabilities to extract accurate and useful representations from the non-Euclidean domain~\cite{kipf2016semi, xu2018powerful}.
With the blossom of more diverse and sophisticated GNN models, the demand for efficient GNN computing is ever-increasing.
The multifarious nature of GNN applications poses great challenges for scalability and efficiency in GNN execution, hence more attention has been drawn to GNN acceleration through software/hardware co-optimization approaches~\cite{abadal2021computing}.

\begin{figure}[t]
\centering
\includegraphics[width=0.47\textwidth]{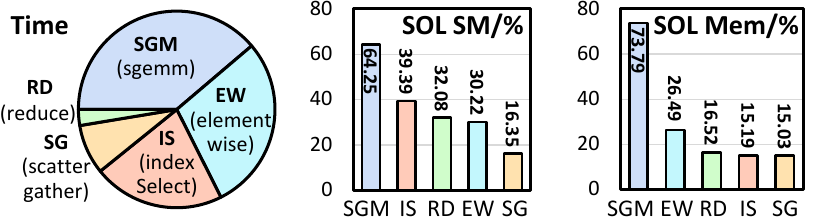}
\caption{Breakdown of a single GCN layer inference on Yelp dataset. 
The Speed Of Light (SOL) metric denotes how close each unit is to its maximum theoretical throughput. SM denotes Streaming Multiprocessor.}
\label{fig:breakdown}
\end{figure}

In general, GNN acceleration focuses on two major phases during GNN inference: \textit{Aggregation} and \textit{Combination}. 
\textit{Aggregation} updates the vertex or edge feature vectors by aggregating features in the neighborhood,
and \textit{Combination} usually puts features through a multi-layer perceptron.
As shown in Fig.~\ref{fig:breakdown}, inference of a typical GNN layer can be broken down into execution of several key kernels.
Aside from the \textit{sgemm} kernel executed during \textit{Combination}, kernels utilized for implementing \textit{Aggregation} (as detailed in Sec. \ref{sec:result}) are also vital regarding their computation and memory requirements.
Moreover, compared with the well-studied matrix operations, the distinctive graph-related nature of \textit{Aggregation} poses additional challenges for its efficient implementation.

To this end, several programming abstractions have been proposed as high-level constructs to streamline the implementation of GNNs, especially for the \textit{Aggregation} phase~\cite{ma2019neugraph, chen2020fusegnn, geng2021gcn, zhang2022understanding}.
Although these abstractions are adopted by most GNN acceleration studies, the choice of a specific abstraction is often determined by pre-configured settings in existing GNN frameworks, and only very few of them critically assess the efficacy of the selected abstraction.
The absence of comparison among these abstractions under a fair evaluation setting also makes analyzing them a critical task for efficient implementations.

In this letter, we aim to carry out a comprehensive evaluation across different programming abstractions for GNN \textit{Aggregation}.
We first summarize the abstractions utilized by existing GNN systems, and taxonomize them along the dimension of data organization and propagation method.
To better understand the execution patterns of each abstraction, we provide characterization on several representative GNN models and datasets from various applications.
We observe that there is no all-time superior solution, for the computation and memory behaviors of GNN programming abstractions are highly dependent on both computation platforms and input graph topologies.
Our evaluation suggests that versatility in programming abstraction selection is needed in future GNN acceleration research.

%% file: main_text/2-Preliminaries.tex
\section{Preliminaries}\label{sec:preliminaries}

In general, the execution of a GNN layer follows a Message Passing paradigm.
For a graph $G = (V,E)$, where $V$ and $E$ denote the set of vertices and edges respectively, the inference of a GNN layer can be formulated as:
\begin{equation}\label{eq:gnn}
    \bm{h}_u^{k}=Comb \left(Aggr\left(\{\bm{h}_u^{k-1}, \bm{h}_v^{k-1}, \bm{e}_{uv}^{k-1} | v\in N(u)\}\right)\right),
\end{equation}
where $\bm{h}_u^{k}$ and $\bm{e}_{uv}^{k}$ represent the feature vector of vertex $u$ and the edge between $u$ and $v$ at layer $k$ respectively, and $N(u)$ represents the neighborhood of vertex $u$.

For the purpose of unification and efficient computation, various GNN programming abstractions have been proposed. 
One of the most influential abstractions is the Scatter-ApplyEdge-Gather-ApplyVertex with Neural Networks (SAGA-NN) propositioned in~\cite{ma2019neugraph}.
\textit{Scatter} passes the vertex features onto the adjacent edges to construct edge data, while \textit{Gather} collects the features from edges to the destination vertices.
These two fundamental operators in GNN \textit{Aggregation} can be defined as:
\begin{equation}
    \begin{aligned}
        \textbf{Scatter:} &~ \bm{m}_{(u,v)} = \phi \left(\bm{h}_u, \bm{h}_v \right), (u,v) \in E ,\\
        \textbf{Gather:} &~ \bm{h}_v = \varphi \left(\{\bm{m}_{(u,v)}~|~(u,v) \in E\}\right),
    \end{aligned}
\end{equation}
where $\phi(\cdot, \cdot)$ is a binary function that takes the features of the attached vertices, and $\varphi(\cdot)$ is a reduction operation performed iteratively in a vertex neighborhood.
These operators are the basic building-blocks of \textit{Aggregation} phase in mainstream GNNs, and are implicitly utilized in most modern GNN libraries.
Following studies~\cite{chen2020fusegnn, zhang2022understanding, zhou2023ugrapher} have further explored the selection, fusion, and parallelization of operators for different models and scenarios.
Although these studies complement and refine the original SAGA-NN, the area of programming abstraction still remains largely underinvestigated.

Aside from the message passing based abstractions, there has also been an increasing effort on computing GNNs from the matrix perspective, such as Sparse-Dense Matrix Multiplication (\textit{SpMM}). 
Computing \textit{Aggregation} results can then be reformulated as matrix operations between adjacency matrix $A$ and feature map $X$.
For instance, Geng et al. discussed the two typical approaches in~\cite{geng2021gcn}: \textit{Pull} and \textit{Push} based computation methods.
The former aggregates vertex features sequentially, pulling them from the neighborhood of each vertex, while the latter computes the aggregated features simultaneously by broadcasting each vertex feature to its neighbors one by one.

%% file: main_text/3-Methodology.tex
\section{Methodology}\label{sec:method}
We now provide a new taxonomy for the \textit{Aggregation} programming abstractions, and outline our evaluation setup for investigating their association with actual performance.

\subsection{Taxonomy of Programming Abstractions}
\begin{figure}[t]
\centering
\includegraphics[width=0.48\textwidth]{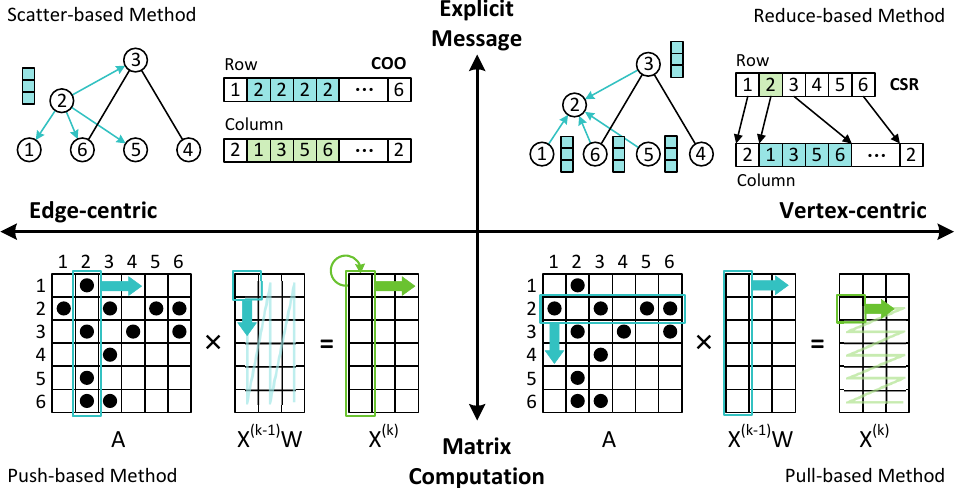}
\caption{Taxonomy of the \textit{Aggregation} programming abstractions.}
\label{fig:aggr_tax}
\end{figure}

Similar to traditional graph analytics, the implementation of \textit{Aggregation} can be classified into either the edge-centric or vertex-centric category.
With the distinction between different data organizations, these categories can be further divided into sub-categories of \textit{Aggregation} with explicit messages or matrix computations.
As shown in Fig.~\ref{fig:aggr_tax}, edge-centric and vertex-centric methods can be categorized as \textit{scatter/push}-based and \textit{reduce/pull}-based methods respectively.

\textbf{Edge-centric} methods take the edges as the topological input and traverse them one by one.
For methods based on explicit messages passing, the input graph structures are usually stored in the Coordinate (COO) format to represent the edge list and sorted by the source vertex ID. 
The feature of each source vertex would be scattered as messages around its neighborhood continuously, and then gathered from the other end once finished.
The \textit{scatter}-based method can be formulated as
\begin{equation}
    \bm{h}_u^k = \varphi (\{\phi (\bm{h}_u^{k-1}, \bm{h}_v^{k-1})~|~(u,v) \in E\}).
\end{equation}
The \textit{scatter}-based abstraction is utilized among several GNN frameworks, such as the scatter-gather method in PyTorch Geometric (PyG)~\cite{Fey/Lenssen/2019}, 
and the Gather-ApplyEdge-Scatter (GAS) abstraction in fuseGNN~\cite{chen2020fusegnn}.
When processing GNNs from the matrix perspective, edge-centric methods take each column of the adjacency matrix $A$ and push the corresponding feature dimension to its neighbors.
One example of the \textit{push}-based method is AWB-GCN~\cite{geng2020awb}, as it utilizes the Compressed Sparse Column (CSC) format for the storage of matrix $A$.

\textbf{Vertex-centric} methods, on the contrary, compute \textit{Aggregation} by sequentially processing each vertex, employing the Compressed Sparse Row (CSR) format to represent the adjacency matrix.
Distinct from edge-centric methods, vertex-centric methods first gather features from the vertex neighborhood and reduce them into new vertex features, then perform a scatter operation to generate new edge features.
The \textit{reduce}-based method can be formulated as
\begin{equation}
    \bm{m}^{k}_{(u,v)} = \phi \left(\varphi \{\bm{m}^{k-1}_{(u,v)}~|~(u,v) \in E\}, \bm{h}^{k-1}_u\right).
\end{equation}
This method is prevalent in both software and hardware GNN acceleration frameworks, such as the SAGA-NN abstraction~\cite{ma2019neugraph}, the earlier versions of Deep Graph Library (DGL)~\cite{wang2019deep}, and HyGCN~\cite{yan2020hygcn}, all of which heavily rely on neighborhood feature reduction.
Similarly, vertex-centric method in matrix form processes $A$ by row, and accesses the features by columns, which is identical to standard matrix multiplication.
The \textit{pull}-based method is also implemented within the majority of GNN frameworks, such as the Memory Efficient Aggregation in PyG, the Matrix View in DGL, and I-GCN~\cite{geng2021gcn}.

The irregularities within the input graph structures raises distinct bottlenecks for both edge- and vertex-centric methods. 
Edge-centric methods, although effective in avoiding vertex feature data locality issues through broadcasting, encounter bottlenecks in memory consumption and workload imbalance. 
Workload balancing techiques, such as the workload distribution autotuning framework in AWB-GCN~\cite{geng2020awb}, have been proposed to mitigate these issues.
Conversely, vertex-centric methods struggle with poor spatial locality but are favored for their lower memory footprint and higher reusability.
Table~\ref{tab:methods-comparison} summarizes the advantages and disadvantages of both approaches.
Beyond these methods, there are also efforts on taxonomizing GNN \textit{Aggregation} by the dimension of \textit{Aggregation} depth, such as the NAU abstraction in FlexGraph~\cite{wang2021flexgraph} and the adaptive optimization policy for aggregation iterations in Policy-GNN~\cite{lai2020policy}.
While they enhance GNN expressiveness and offer more optimization avenues, their computations remain edge- or vertex-centric.
Thus, despite the intricacies of \textit{Aggregation} depth meriting their own discussion, this work primarily focuses on edge- and vertex-centric methods.

\begin{table}[t]
\caption{Comparison between Edge- and Vertex-centric Methods}
\label{tab:methods-comparison}
\centering
\begin{tabular}{ccccc}
\hline
\begin{tabular}[c]{@{}c@{}}Programming\\ Abstraction\end{tabular} &
  \begin{tabular}[c]{@{}c@{}}Reuse\\ of $A$\end{tabular} &
  \begin{tabular}[c]{@{}c@{}}Reuse\\ of $X$\end{tabular} &
  \begin{tabular}[c]{@{}c@{}}Storage of\\ Partial Sum\end{tabular} &
  \begin{tabular}[c]{@{}c@{}}Load\\ Imbalance\end{tabular} \\ \hline
Edge-centric &
  Low &
  High &
  Large &
  High \\
Vertex-centric &
  High &
  Low &
  Small &
  Low \\ \hline
\end{tabular}
\end{table}

\subsection{Evaluation Setup}
\begin{figure}[t]
\centering
\includegraphics[width=0.48\textwidth]{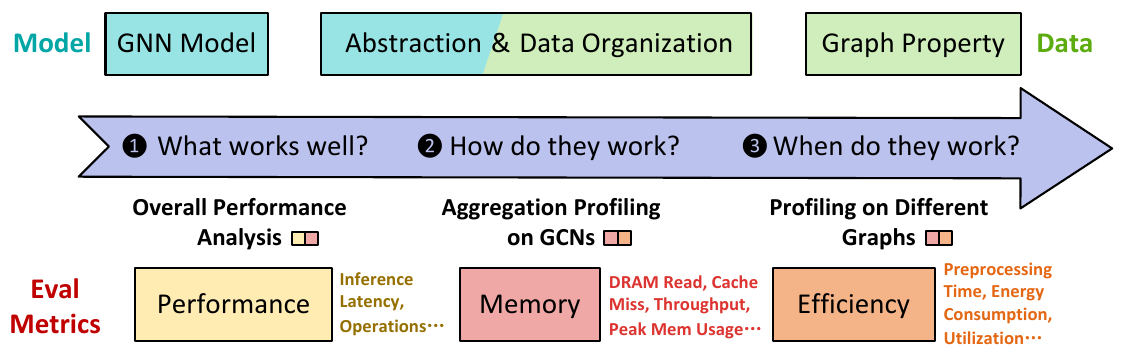}
\caption{Overview of the evaluation settings.}
\label{fig:overview}
\end{figure}

Considering the distinct characteristics of these abstractions, to determine the performance and compatibility of each abstraction with different scenarios requires a comprehensive evaluation.
As shown in Fig.~\ref{fig:overview}, we intend to analyze the characteristics of each abstraction from both model and data perspective.
Hence, we focus on three aspects: \textit{GNN models}, \textit{programming abstractions coupled with data organizations}, and \textit{input graph properties}.
Our evaluation aims to address the following questions:
\begin{dingautolist}{182}
    \item Which programming abstraction obtains the best overall performance?
    \item How does the choice of programming abstraction affect the execution of \textit{Aggregation} phase?
    \item Under what scenario does each programming abstraction perform the best?
\end{dingautolist}

We evaluate these abstractions by studying the inference process of them on the contemporary Intel Xeon E5-2620 and i9-13900 CPUs, and NVIDIA V100, Tesla M40, A100 and the RTX4090 GPUs with CUDA 11.3.
We choose the PyG framework since it supports both vertex-centric and edge-centric programming abstractions.
We mainly focus on three most utilized abstractions: \textit{scatter}-, \textit{reduce}- and \textit{pull}-based methods.
In PyG, these abstractions can be implemented through utilizing \textit{segment\_coo} and \textit{segment\_csr} operations in \textit{pytorch-scatter} package, and \textit{SparseTensor} class from \textit{torch-sparse} package.

To best evaluate the characteristics of each abstraction, we perform experiments on both synthetic and real-world graphs.
The real-world datasets are selected among vertex and graph classification applications, with the number of vertices ranging from 17 to 0.7M, as shown in Table~\ref{tab:dataset}.
For GNNs, we select four representative models: GCN~\cite{kipf2016semi}, GIN~\cite{xu2018powerful}, GAT~\cite{velickovic2017graph} and PDN~\cite{rozemberczki2021pathfinder}.
The former three are among the most studied and implemented GNN models, while the last one is chosen because of its exploitation of edge features.
We only run experiments with scatter-based and reduce-based methods for GAT, since the attention mechanism is not natively supported by the \textit{SparseTensor} class.

\begin{table}[ht]
\centering
\caption{Real-world Graph Datasets}
\resizebox{0.48\textwidth}{!}{
\begin{tabular}{ccccccc}
\hline
Application                            & Dataset & Graph \# & \begin{tabular}[c]{@{}c@{}}Avg. \\ Vertex \#\end{tabular} & \begin{tabular}[c]{@{}c@{}}Avg. \\ Edge \#\end{tabular} & \begin{tabular}[c]{@{}c@{}}Feature\\ Length\end{tabular} & Classes \\ \hline
\multirow{5}{*}{\begin{tabular}[c]{@{}c@{}}Vertex \\ Classification \end{tabular}} & Cora (CR)    & 1        & 2708                                                      & 10556                                                   & 1433                                                     & 7       \\
                                       & Pubmed (PB)  & 1        & 19717                                                     & 88648                                                   & 500                                                      & 3       \\
                                       & Flickr (FR)  & 1        & 89250                                                     & 899756                                                  & 500                                                      & 7       \\
                                       & Yelp (YP)    & 1        & 716847                                                    & 13954819                                                & 300                                                      & 100     \\
                                       & Reddit (RD)  & 1        & 232965                                                    & 23213838                                                & 602                                                      & 41      \\ \hline
\multirow{2}{*}{\begin{tabular}[c]{@{}c@{}}Graph \\ Classification \end{tabular}}  & PPI     & 20       & 2245.3                                                    & 61318.4                                                 & 50                                                       & 121     \\
                                       & MUTAG (MT)   & 188      & 17.9                                                      & 39.6                                                    & 7                                                        & 2       \\ \hline
\end{tabular}}
\label{tab:dataset}
\end{table}

%% file: main_text/4-Result-Analysis.tex
\section{Observation and Analysis}\label{sec:result}

\subsection{Analysis on Overall Performance}
To analyze the overall performance of each programming abstraction, we run the forward pass of a single GNN layer on real-world graph learning datasets. 
In each run, we perform the inference process 300 times and take their average result.
The input feature length is the native dimension length of each dataset, and the output dimension is set to 8 for simplicity.

\begin{figure}[t]
\centering
\includegraphics[width=0.48\textwidth]{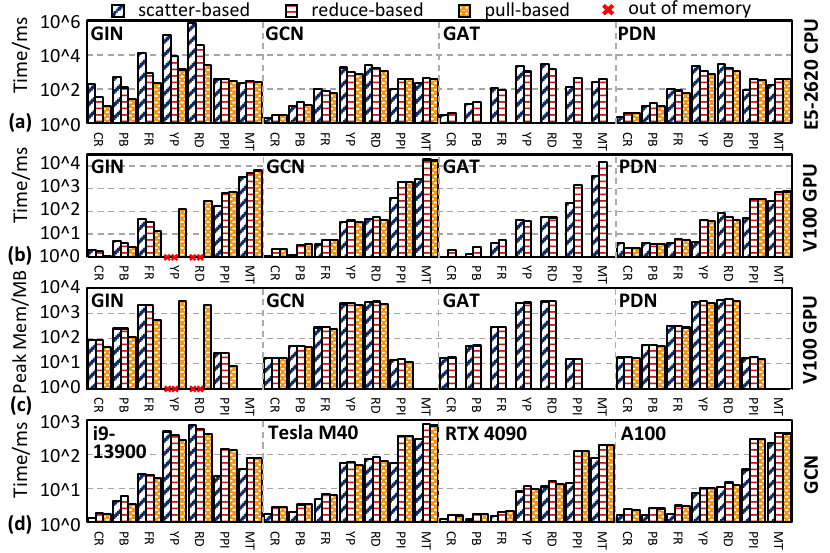}
\caption{The inference latency and peak memory usage of the selected GNNs. Here we use the logarithm of results for clearer illustration.}
\label{fig:latency&peakmem}
\end{figure}

\textbf{Inference Latency Analysis.} 
Fig.~\ref{fig:latency&peakmem} (a) and (b) illustrate the total execution latency on the Intel E5-2620 CPU and V100 GPU.
The inference time for all abstractions generally increases with the scale of datasets.
On large scale vertex classification datasets, \textit{pull}-based method usually procures the lowest latency, 
while the \textit{scatter}-based method acquires relatively better results on smaller graphs.
Note that due to GIN executes \textit{Aggregation} before \textit{Combination}, 
its inference latency is significantly influenced by the input vertex feature length.
Consequently, \textit{pull}--based method is optimal for all vertex classification datasets with GIN.

\textbf{Peak Memory Usage.}
Fig.~\ref{fig:latency&peakmem} (c) compares the peak memory usage on the V100 GPU, which also scales up as the graph size increases.
On smaller graphs, the peak memory usage of these methods are roughly the same, and the advantage of \textit{pull}-based method emerges as the graph size increases.
Moreover, due to the limited VRAM resources, processing large graphs with high feature dimensions is likely to encounter out of memory errors.

\textbf{Platform-wise Performance Comparison.}
Fig.~\ref{fig:latency&peakmem} (d) depicts the execution latency of the GCN model on the i9-13900 CPU, as well as the Tesla M40, RTX 4090 and A100 GPUs.
Results on i9-13900 CPU and Tesla M40 GPU align with our prior findings.
However, on higher-end platforms like the RTX 4090 and A100 GPUs, \textit{scatter}-based method remains optimal for large-scale datasets.
This is attributed to the edge list format utilize by \textit{scatter}-based method, which offers a more extensive avenue for parallelism than the compressed adjacency matrix format. 
Consequently, on these higher-end platforms, benefits of parallelization overshadow the overhead from message construction and storage.
It is also noteworthy that for graph classification datasets, latency on CPUs is consistently lower than on GPUs.
This discrepancy stems from the repeated preprocessing and data transfer time required for each graph, in contrast to vertex classification where a single graph is processed only once.

\subsection{Aggregation Profiling on GCN}
To better understand the execution of GNN \textit{Aggregation}, we dissect the \textit{Aggregation} phase of GCN on four vertex classification datasets.
Using Linux Perf on the E5-2620 CPU and NVIDIA Nsight Compute on the V100 GPU as profiling tools, we report the key findings as follows.
Notably, our GPU profiling involves a fine-grained kernel-level inspection.
These kernels include \textit{indexSelect},  \textit{unrolled\_elementwise}, \textit{scatter\_gather\_elementwise} (\textit{scatter}), \textit{segment\_csr\_broadcast} (\textit{reduce}) and \textit{spmm} (\textit{pull}).

\begin{figure}[tbp]
\centering
\includegraphics[width=0.48\textwidth]{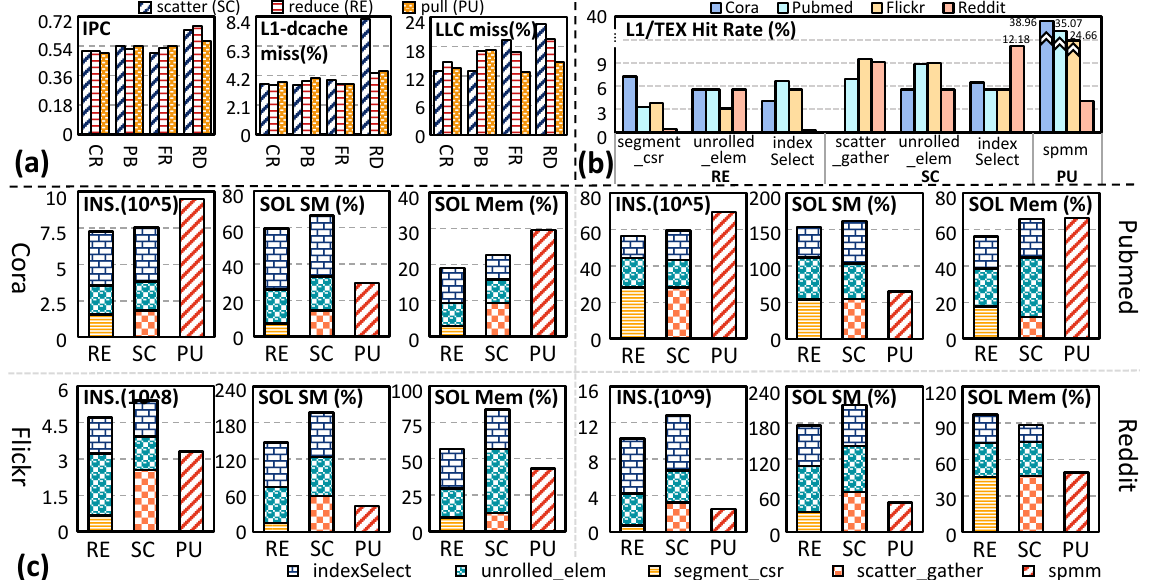}
\caption{Profiling results of GCN \textit{Aggregation}: (a) Results on Intel E5-2620 CPU, (b) L1/TEX hit rate on V100 GPU, and (c) Kernel level profiling on V100 GPU.}
\label{fig:agg_prof}
\end{figure}

\textbf{Computational Efficiency.}
As shown in Fig.~\ref{fig:agg_prof} (a) and (c), the number of execution instructions on both CPU and GPU scale with the dataset size, consistent to the overall performance analysis results.
However, compared to matrix-based methods, message-based methods exhibit a steeper increase in total instructions as graph size grows, leading to a relatively higher IPC on large graphs.
In addition, the computation throughput and total instructions on GPU indicate that although \textit{indexSelect} is the most computation intensive kernel in message-based methods, the difference between \textit{scatter}- and \textit{reduce}-based methods is attributed to the distinction between \textit{scatter\_gather} and \textit{segment\_csr}.
Hence, the latter obtains slightly better performance over the former on larger graphs.

\textbf{Cache and Memory Utilization.}
As shown in Fig.~\ref{fig:agg_prof} (a), both the L1 and LLC cache load miss on CPU conform with previous findings:
the cache miss ratio for message-based methods rises as dataset scale increases, especially for \textit{scatter}-based method.
In contrast, the cache access patterns on GPU, depicted in Fig.~\ref{fig:agg_prof} (b), deviate from previous computation patterns.
Notably, \textit{spmm} is more sensitive to dataset scale, as its cache hit rate drops drastically when the graph size increases.
While both \textit{scatter}- and \textit{reduce}-based methods utilize \textit{indexSelect}, there exist disparities in their respective cache behaviors.
Specifically, as the dataset size increases, the hit rate of \textit{indexSelect} rises in \textit{scatter}-based but falls in \textit{reduce}-based method.
This can be attributed to the fact that the former only collects the indices of the neighboring vertices before message scattering, while the latter also gathers vertex features.

\begin{figure}[t]
\centering
\includegraphics[width=0.49\textwidth]{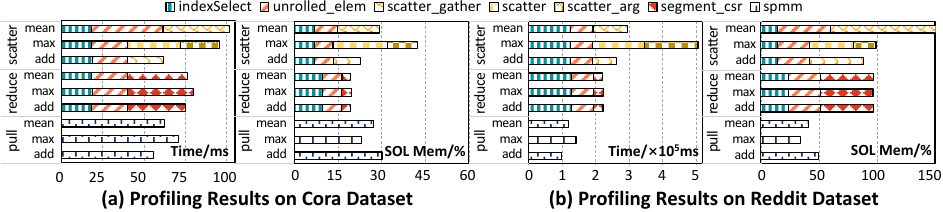}
\caption{Profiling results with different \textit{Aggregation} operators on V100 GPU.}
\label{fig:prof_op}
\end{figure}

\textbf{Aggregation Operators.}
We further analyze the effect of \textit{Aggregation} operators by profiling GCN \textit{Aggregation} with three operators: \textit{add}, \textit{max} and \textit{mean}.
As shown in Fig.~\ref{fig:prof_op}, \textit{reduce}-based method performs most consistently across all three operators, while the results with \textit{scatter}-based method vary the most.
Compared to \textit{add} and \textit{mean}, \textit{max} operator requires comparison among the vertex neighborhood, which can impede the memory utilization and contribute to longer processing time.

\subsection{Profiling with Different Graph Properties}
To pinpoint the ideal scenario for each abstraction, we conduct additional experiments on the V100 GPU, with synthesized datasets featuring diverse structural properties.
We use \textit{igraph} and \textit{NetworkX} to generate a series of random graph structures with 10K vertices and varying structural properties, including density, power-law exponent and clustering coefficient. 

\begin{figure}[t]
\centering
\includegraphics[width=0.49\textwidth]{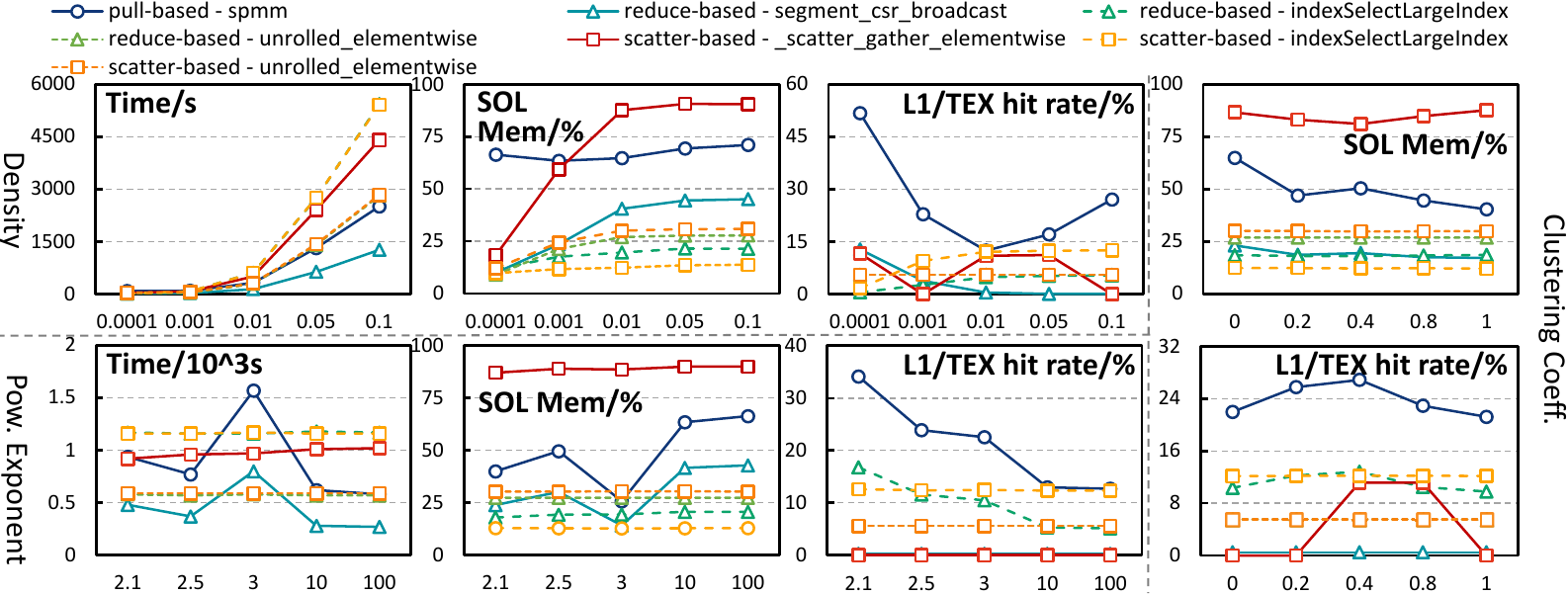}
\caption{Profiling results on synthesized graph datasets with different properties. }
\label{fig:graphprof}
\end{figure}

\textbf{Density.}
For a graph with fixed number of vertices, its density is directly associated with computation and memory requirements, as it depicts the number of edges within the graph.
As shown in Fig.~\ref{fig:graphprof}, the growth of memory utilization in message-based methods decelerates significantly after graph density reaches 0.01, due to computational resource constraints.
This density also marks a turning point for L1 cache hit rate with the \textit{spmm} kernel, which initially declines sharply before gradually recovering.
The impact of limited cache capacity on spatial locality first becomes more pronounced with increasing density, but is gradually mitigated by increased reuse distance.

\textbf{Other Graph Properties.}
Although the amount of computation is predetermined by the number of vertices and edges, the edge distribution still has a significant impact on inference efficiency and memory utilization.
The power-law exponent depicts the skewness of vertex degree distribution.
As shown in Fig.~\ref{fig:graphprof}, for vertex-centric methods, smoother degree distribution results in improved latency and memory utilization, while also leading to lower cache hit rate due to fewer data reuse opportunities.
Note that the spike at 3 is caused by the finite size correction in \textit{igraph} for exponents less than 3.
Moreover, the impact of clustering on computation and memory efficiency is found to be minor for less memory intensive kernels.

%% file: main_text/5-Discussion.tex
\section{Discussion}\label{sec:discussion}

In this section, we present some insights inspired by our evaluation results.
\ding{182}~\textbf{Abstraction selection}:
Our findings reveal a clear pattern in the relationship between graph size and \textit{Aggregation} abstraction efficiency.
Specifically, \textit{scatter}-based method favors smaller graphs, whereas \textit{pull}-based method suits larger ones.
\ding{183}~\textbf{Hardware adaptability}:
Data organizations associated with these abstractions significantly  influence the performance across platforms.
While compressed matrix format is generally more adaptable and efficient, edge list format thrives on higher-end platforms.
Moreover, when handling multiple graphs, as in the case of graph classification, the overheads of preprocessing and data transfer to GPUs merit considerable attention.
\ding{184}~\textbf{Structural impact}: Beyond size, the intrinsic properties of input graphs also greatly affect \textit{Aggregation}, with density emerging as a pivotal factor shaping performance. 
Furthermore, the skewness in vertex degree distribution appears to favor vertex-centric methods.

%% file: main_text/6-Conclusion.tex
\section{Conclusion}\label{sec:conclusion}

In this paper, we classify existing GNN \textit{Aggregation} programming abstractions, and characterize these abstractions by performing comprehensive evaluations on various platforms.
Our analysis suggests that the performances of GNN abstractions are highly dependent on hardware platforms and input graph structures, and each abstraction has its own suitable application scenarios.
We believe that our observations can aid programmers and researchers in distinguishing each abstraction, and provide guidance for future research on GNN acceleration.

%% file: main.bbl
\begin{thebibliography}{10}
\providecommand{\url}[1]{#1}
\csname url@samestyle\endcsname
\providecommand{\newblock}{\relax}
\providecommand{\bibinfo}[2]{#2}
\providecommand{\BIBentrySTDinterwordspacing}{\spaceskip=0pt\relax}
\providecommand{\BIBentryALTinterwordstretchfactor}{4}
\providecommand{\BIBentryALTinterwordspacing}{\spaceskip=\fontdimen2\font plus
\BIBentryALTinterwordstretchfactor\fontdimen3\font minus \fontdimen4\font\relax}
\providecommand{\BIBforeignlanguage}[2]{{%
\expandafter\ifx\csname l@#1\endcsname\relax
\typeout{** WARNING: IEEEtran.bst: No hyphenation pattern has been}%
\typeout{** loaded for the language `#1'. Using the pattern for}%
\typeout{** the default language instead.}%
\else
\language=\csname l@#1\endcsname
\fi
#2}}
\providecommand{\BIBdecl}{\relax}
\BIBdecl

\bibitem{kipf2016semi}
M.~Welling \emph{et~al.}, ``Semi-supervised classification with graph convolutional networks,'' in \emph{Proc. of ICLR}, 2016.

\bibitem{xu2018powerful}
K.~Xu \emph{et~al.}, ``How powerful are graph neural networks?'' in \emph{Proc. of ICML}, 2019.

\bibitem{abadal2021computing}
S.~Abadal \emph{et~al.}, ``Computing graph neural networks: A survey from algorithms to accelerators,'' \emph{ACM CSUR}, pp. 1--38, 2021.

\bibitem{ma2019neugraph}
L.~Ma \emph{et~al.}, ``{NeuGraph}: Parallel deep neural network computation on large graphs,'' in \emph{Proc. of USENIX ATC}, 2019, pp. 443--458.

\bibitem{chen2020fusegnn}
Z.~Chen \emph{et~al.}, ``{fuseGNN}: accelerating graph convolutional neural network training on gpgpu,'' in \emph{Proc. of ICCAD}, 2020, pp. 1--9.

\bibitem{geng2021gcn}
T.~Geng \emph{et~al.}, ``{I-GCN}: A graph convolutional network accelerator with runtime locality enhancement through islandization,'' in \emph{Proc. of MICRO}, 2021, pp. 1051--1063.

\bibitem{zhang2022understanding}
H.~Zhang \emph{et~al.}, ``Understanding gnn computational graph: A coordinated computation, io, and memory perspective,'' in \emph{Proc. of MLSys}, 2022.

\bibitem{zhou2023ugrapher}
Y.~Zhou \emph{et~al.}, ``{uGrapher}: High-performance graph operator computation via unified abstraction for graph neural networks,'' in \emph{Proc. of ACM ASPLOS}, 2023, pp. 878--891.

\bibitem{Fey/Lenssen/2019}
M.~Fey \emph{et~al.}, ``Fast graph representation learning with {PyTorch Geometric},'' in \emph{ICLR Workshop}, 2019.

\bibitem{geng2020awb}
T.~Geng \emph{et~al.}, ``{AWB-GCN}: A graph convolutional network accelerator with runtime workload rebalancing,'' in \emph{Proc. of MICRO}, 2020, pp. 922--936.

\bibitem{wang2019deep}
M.~Y. Wang, ``Deep graph library: Towards efficient and scalable deep learning on graphs,'' in \emph{ICLR workshop}, 2019.

\bibitem{yan2020hygcn}
M.~Yan \emph{et~al.}, ``{HyGCN}: A {GCN} accelerator with hybrid architecture,'' in \emph{Proc. of IEEE HPCA}, 2020, pp. 15--29.

\bibitem{wang2021flexgraph}
L.~Wang \emph{et~al.}, ``Flexgraph: a flexible and efficient distributed framework for gnn training,'' in \emph{Proc. of EuroSys}, 2021, pp. 67--82.

\bibitem{lai2020policy}
K.-H. Lai \emph{et~al.}, ``Policy-gnn: Aggregation optimization for graph neural networks,'' in \emph{Proc. of SIGKDD}, 2020, pp. 461--471.

\bibitem{velickovic2017graph}
P.~Velickovic \emph{et~al.}, ``Graph attention networks,'' \emph{arXiv:1710.10903}, 2017.

\bibitem{rozemberczki2021pathfinder}
B.~Rozemberczki \emph{et~al.}, ``Pathfinder discovery networks for neural message passing,'' in \emph{Proc. of the Web Conference}, 2021, pp. 2547--2558.

\end{thebibliography}
